\documentclass[conference, onecolumn]{IEEEtran}
\IEEEoverridecommandlockouts
\usepackage{amsmath,amssymb,amsfonts}
\usepackage{algorithmic}
\usepackage{graphicx}
\usepackage{textcomp}
\usepackage{xcolor}

\usepackage[numbers,sort&compress]{natbib}%步骤1
\usepackage{array}
\usepackage{booktabs}
\usepackage{multirow}
\usepackage[normalem]{ulem}
\useunder{\uline}{\ul}{}

\def\BibTeX{{\rm B\kern-.05em{\sc i\kern-.025em b}\kern-.08em
    T\kern-.1667em\lower.7ex\hbox{E}\kern-.125emX}}
\begin{document}

\title{An Efficient and Scalable Graph Condensation with Structure-Preserving\\
}

\author{
\IEEEauthorblockN{Yulin Hu}
\IEEEauthorblockA{\textit{Southwest University} \\
Chongqing, China \\
hylinserein@email.swu.edu.cn}
\and
\IEEEauthorblockN{Fuyan Ou}
\IEEEauthorblockA{\textit{Southwest University} \\
Chongqing, China \\
omaksimovbe@gmail.com}
\and
\IEEEauthorblockN{Ye Yuan}
\IEEEauthorblockA{\textit{Southwest University} \\
Chongqing, China \\
yuanyekl@swu.edu.cn}
}
% \author{\IEEEauthorblockN{Anonymous Authors}}

\maketitle

\begin{abstract}
Graph condensation (GC) is pivotal for enabling Graph Neural Networks (GNNs) deployment in resource-constrained scenarios by compressing large-scale graphs into compact synthetic counterparts. Existing GC methods commonly suffer from computational inefficiency due to coupled optimization as well as encountering poor generalization across GNN architectures. To address these challenges, this study proposes an \underline{E}fficient and \underline{S}calable \underline{G}raph \underline{C}ondensation with \underline{S}tructure-\underline{P}reserving (SP-ESGC), which possesses a decoupled design that separates node condensation from graph structure generation. Specifically, it first employs heat kernel feature propagation to generate node representation via spectral graph theory-inspired diffusion. Further, a novel hybrid clustering strategy is designed to extracts discriminative intra-class centroids from the node representation. Finally, a pre-trained edge predictor infers transferable structural patterns from the original graph, ensuring accurate synthetic graph generation. Extensive experiments on real-world graph datasets demonstrate that the proposed SP-ESGC implementes a precise GC with significantly high computational efficiency. Moreover, SP-ESGC also generalizes well across diverse GNN architectures.
\end{abstract}

\begin{IEEEkeywords}
Graph Condensation, Graph Neural Networks, Graph Representation Learning, Efficiency
\end{IEEEkeywords}

\section{INTRODUCTION}
Graph-structured data is fundamental to many applications, including social networks \cite{r1,a20,a21,a22}, molecular modeling \cite{r2,a16,a17,a18,a19}, and recommender systems \cite{r3,a7,a8}. Graph neural networks (GNNs) \cite{r4,a4,a5} can effectively learn from such data, but training GNNs on large graphs is increasingly expensive, limiting their use in resource-constrained scenarios, such as neural architecture search and continual learning\cite{r5,a6,a9}. Graph condensation has emerged as a promising solution by condensing large graphs into small synthetic ones while preserving downstream performance\cite{r6,a10,a11}. Inspired by dataset condensation in the image domain \cite{r7}, early methods typically retain information through gradient matching, distribution matching, or trajectory matching.

Despite improved condensation quality, these methods still have several problems. A major issue lies in their optimization design. Most methods tightly couple the optimization of node features, graph structures, and GNN parameters\cite{a1,a2,a3}. It often requires bi-level or even tri-level optimization, making the condensation process slow and difficult to scale, especially for large-scale graphs. Another issue is the dependence on specific GNN architectures. In many methods, the condensation objective is closely tied to the relay model. This tight coupling not only weakens generalization across different architectures but also leads to unstable performance when applied in practical settings. Finally, preserving graph structural information remains a challenge.  Although some methods\cite{SGDD} attempt to introduce structural constraints, they often do so at the cost of increased computational overhead or reduced flexibility.

Recent work has explored more efficient condensation schemes through simplified optimization objectives. However, several challenges remain unresolved\cite{r8,a23}, including how to obtain stable node representations with lightweight optimization\cite{r12}, how to preserve class-level discriminability under large node reduction, and how to efficiently generate the topology of the synthetic graph\cite{a24,a25}. To address these challenges, we propose SP-ESGC, an efficient graph condensation framework with a decoupled design. We separate node feature condensation from graph structure generation to improve efficiency and stability. Specifically, we employ heat kernel feature propagation to obtain node representations, followed by a clustering-based condensation strategy that integrates low-rank approximation and kernel feature expansion to extract representative class centroids. Finally, we generate the synthetic graph structure using an edge predictor pre-trained on the original graph.

Our main contributions can be summarized as follows:
\begin{itemize}
\item We propose a graph condensation framework that avoids costly bi-level optimization and repeated GNN training.
\item We design a feature-based graph structure generation mechanism that enables efficient and scalable topology construction.
\item Our method significantly reduces the computational cost without compromising the performance of downstream tasks, making it suitable for large-scale graph scenarios.
\end{itemize}

\section{RELATED WORK}

\textbf{Gradient Matching.} Gradient matching is an early and influential paradigm in graph condensation\cite{r9,a26,a27,a28}, formulating condensation as a bi-level optimization problem by aligning GNN parameter gradients between original and condensed graphs. GCond \cite{GCond} enforces gradient consistency to approximate original training dynamics and achieves high condensation ratios. DosCond \cite{DosCond} simplifies this framework via one-step gradient matching, eliminating relay updates and hyperparameter tuning. SGDD \cite{SGDD} further incorporates structural information to mitigate spectral mismatches. Although these methods are effective, gradient matching is still computationally expensive and sensitive to GNN architectures, which limits its scalability.

\textbf{Trajectory Matching.} Trajectory matching extends gradient matching by aligning full training dynamics rather than gradients. SFGC \cite{SFGC} argues that gradient matching cannot capture complex GNN learning behavior, it matches multi-step training trajectories to encode richer dynamics from the original graph. GEOM \cite{GEOM} further observes that limited supervision in trajectory matching leads to performance gaps. Despite improved condensation quality, trajectory matching incurs substantial computational and storage overhead due to teacher model pretraining and trajectory storage\cite{a29}, limiting scalability to large graphs and resource-constrained settings.

\textbf{Kernel Ridge Regression.} Kernel-based methods\cite{r11} reformulate graph condensation to avoid expensive GNN training. GC-SNTK \cite{GC-SNTK} models GNN behavior using graph neural tangent kernels and casts condensation as a kernel ridge regression problem, eliminating bi-level optimization and repeated GNN training. KiDD \cite{KiDD} extends this formulation to graph classification by removing nonlinear activations in the kernel\cite{a12,a13,a14,a15}, simplifying matrix multiplications during condensation and achieving higher accuracy than gradient matching. While these methods improve efficiency and stability, their performance depends on kernel expressiveness and may be limited in capturing complex nonlinear patterns in real-world graphs.

\textbf{Distribution Matching.} Distribution matching\cite{r10} aims to align statistical distributions between the original and condensed graphs. GCDM \cite{GCDM} models node receptive fields and represents the original graph as a collection of receptive field distributions. Consistency\cite{a30} between the original and synthetic graphs is enforced using Maximum Mean Discrepancy (MMD), which reduces dependence on specific GNN architectures and improves generalization across models\cite{a31}. SimGC \cite{SimGC} further incorporates pre-trained GNNs to align node representation variations and enhance performance. However, as these methods rely on static distribution alignment\cite{a32}, they often struggle to preserve structural information and dynamic training behavior, particularly under large condensation ratios.

\section{METHODOLOGY}
\begin{figure*}[htbp]
\centering
\includegraphics[width=0.9\textwidth]{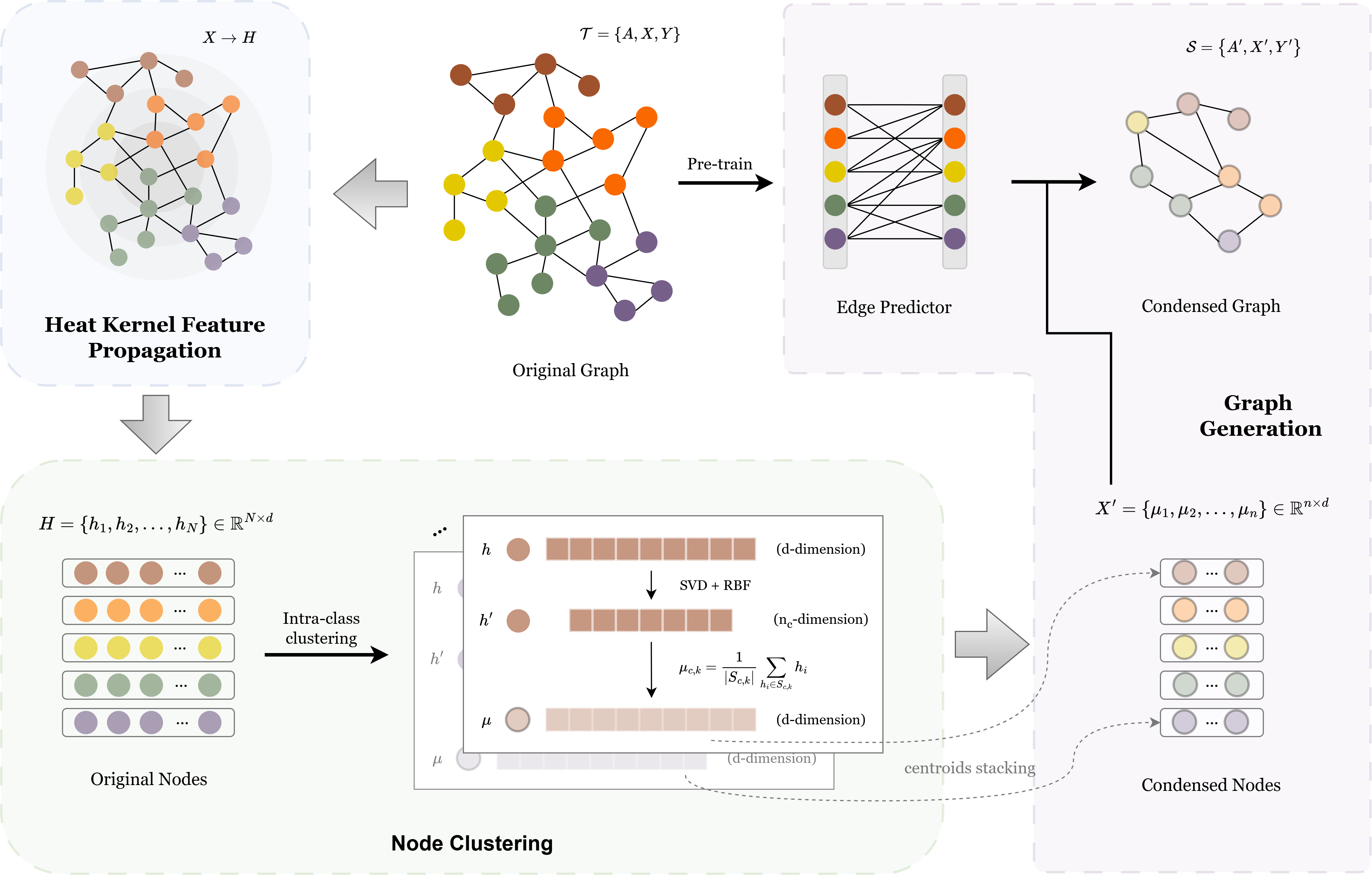}
\caption{A schematic diagram of the proposed SP-ESGC framework.}
\label{fig1}
\end{figure*}

The objective of graph condensation is to construct a synthetic graph from a large-scale original graph, while retaining the information that is essential for downstream tasks. By reducing the size of the training data, graph condensation aims to significantly lower computational and storage costs without sacrificing model performance. In this section, we will introduce the graph condensation framework SP-ESGC that we proposed, as illustrated in Fig.\ref{fig1}. Consider that we have a graph set $\mathcal{T}=\left\{A,X,Y\right\}$, where $A\in\mathbb{R}^{N\times N}$ is the adjacency matrix, $N$ is the number of nodes, $X\in\mathbb{R}^{N\times d}$ is the d-dimensional node feature matrix and $Y\in\{0,\ldots,C-1\}^{N}$ represents the node labels.Graph condensation will generate a small synthetic graph $\mathcal{S}=\left\{A',X',Y'\right\}$, where $A'\in\mathbb{R}^{n\times n}$, $X'\in\mathbb{R}^{n\times d}$, $Y'\in\{0,\ldots,C-1\}^{n}$, such that the GNN trained on $\mathcal{S}$ can achieve comparable performance to the GNN trained on $\mathcal{T}$.

\subsection{Heat Kernel Feature Propagation}\label{AA}
In graph-structured data, node features are often affected by noise and inconsistencies in graph structure or limited local neighborhoods can make it difficult for learning models to capture global information\cite{a33,a34}. To further enhance the stability of feature representations and improve structural coherence, we introduces heat kernel feature propagation, a mechanism inspired by spectral graph theory that provides a smooth, stable and controllable way to diffuse node features across the graph.

The theoretical foundation of heat kernel feature propagation originates from the classical heat equation defined on Euclidean spaces\cite{r13}. In particular, the heat equation describes how heat diffuses through a medium over time:
\begin{equation}\frac{\partial u}{\partial t}=\Delta u,\end{equation}
where $\Delta$ is the continuous Laplace operator. However, graph is not continuous space, its nodes do not have continuous coordinates but only discrete edge relationships, so the continuous heat equation cannot be applied directly.

To adapt the heat diffusion model to graph, we convert the continuous heat equation into a discrete form on graphs\cite{r14,a35}. We first construct the symmetric normalized adjacency matrix $S=D^{-\frac{1}{2}}AD^{-\frac{1}{2}}$, where $A$ is the adjacency matrix and $D$ is the degree matrix. Based on this, the symmetric normalized graph Laplacian is defined as $L=I-S$. By replacing the continuous Laplace operator $\Delta$ with the graph Laplacian $L$, the heat equation on graphs can be written as:
\begin{equation}\frac{\partial{X}(t)}{\partial t}=-L{X}(t),{\quad}with{\ } initial\quad{X}(0)=X,\end{equation}
where $X(t)$ denotes the node features evolving over time $t$. This equation admits an analytical solution:
\begin{equation}X(t)=e^{-tL}X,\end{equation}
where $e^{-tL}$ is known as the heat kernel on graphs. Directly computing $e^{-tL}$ on large-scale graphs is computationally expensive. To maintain scalability, we adopt a truncated Taylor series approximation:
\begin{equation}e^{-tL}\approx\sum_{k=0}^K\frac{(-t)^k}{k!}L^k,\end{equation}
where $K$ is the truncation coefficient. By retaining the first $K$ terms of the infinite series, we obtain an approximation that balances accuracy and computational cost\cite{r15}. Therefore, we ultimately obtain the node representation matrix after heat kernel feature propagation:
\begin{equation}H=\mathbf{X}(t)\approx\sum_{k=0}^K\frac{(-tL)^k}{k!}X.\end{equation}

$H$ not only has a smoother representation, but also effectively integrates the local and global information of the graph\cite{a36}, providing more expressive and stable input features for the subsequent graph condensation module.

\subsection{Node Clustering}
In graph condensation tasks, we aim to condense the set of nodes belonging to each class into a small number of representative intra-class centroids while preserving discriminative information among different classes\cite{a37,a38}. These centroids serve as compact representations of intra-class information for subsequent tasks\cite{a39,a40}, and their quality directly affects the overall performance of downstream models trained on the condensed graph\cite{CGC}. To achieve this, we propose a hybrid clustering method for condensing node representations, which integrates low-rank approximation, randomized kernel feature expansion and spectral space clustering to obtain a robust and discriminative set of intra-class centroids\cite{a46}.

Given node representation matrix $H=\{h_1,h_2,\ldots,h_N\}\in\mathbb{R}^{N\times d}$ and the corresponding class labels $y\in\{0,\ldots,C-1\}^{N}$, the objective is to generate $n_{c}$ representative node centroids within each class $c$ from the set of node representations $H_c=\{h_i\mid y_i=c\}\in\mathbb{R}^{N_c\times d}$. These centroids form a class representation matrix 
\begin{equation}H'_c=\{\mu_1,\mu_2,\ldots,\mu_{n_c}\},\end{equation}
where each centroid $\mu_j\in\mathbb{R}^d$ approximates the representation of samples within the same cluster and serves as a representative of the corresponding class\cite{a41,a42}.

For each class $c$, we extract the corresponding feature matrix $H_c$. To suppress mean shift, we first apply mean-centering to the features. We then perform truncated singular value decomposition (SVD)\cite{r16,a43} to extract a low-rank information with rank is aligned with the target number of intra-class clusters\cite{a44}. This step captures the directions of greatest variance within the class. By discarding components associated with smaller singular values, it provides a low-noise and low-rank basis space $U_{c}$ for kernel expansion. Since node representations often lie on complex nonlinear structures, relying solely on linearly reduced features fails to achieve effective separation. To enhance the nonlinear discriminability of the features, we apply Random Fourier Features(RFF) to $U_{c}$, approximating the mapping of the Radial Basis Function(RBF) kernel:
\begin{equation}Z_c=\phi_{\text{RBF}}(U_c)\in\mathbb{R}^{N_c\times m},\end{equation}
where $m$ is the random feature dimension used to approximate the RBF kernel, which represents the length of the feature vector generated for each sample through RFF\cite{r17}. According to Bochner's theorem, any continuous and positive-definite function can be expressed as the Fourier transform of some probability density function. Therefore:
\begin{equation}\phi(x)=\sqrt{\frac{2}{m}}\cos(\omega^\top x+b).\end{equation}

This process embeds node representations into a more discriminative feature space without explicitly constructing the kernel matrix. The resulting $Z_{c}$ is then further condensed in feature dimension to extract the most stable cluster information, ensuring that the retained features better capture the variation directions of local clusters. This yields the spectral embedding space $Q_c\in\mathbb{R}^{N_c\times n_c}$. Clustering is then performed within each class in $Q_{c}$\cite{a45}. The intra-class centroids are computed in the original feature space by using the cluster labels $\left\{S_{c,k}\right\}$:
\begin{equation}\mu_{c,k}=\frac{1}{|S_{c,k}|}\sum_{h_i\in S_{c,k}}h_i.\end{equation}

Finally, by sequentially stacking the centroids of all classes, we obtain the node representations of the condensed graph:
\begin{equation}X'=\bigcup_{c=0}^{C-1}\bigcup_{k=1}^{n_c}\mu_{c,k}\in\mathbb{R}^{n\times d}.\end{equation}

\subsection{Graph Generation}
We construct the condensed graph structure using a feature-based edge predictor learned from the original graph\cite{a47,a48,a49}. This approach consists of two stages: (i) pretraining a parameterized edge predictor on the original graph, (ii) performing all-pairs inference with the pre-trained model on the synthetic node features to construct the final condensed graph structure.

Our goal is given a set of synthetic node features $X'$ to generate a condensed graph $\mathcal{S}=\left\{A',X',Y'\right\}$ whose structure reflects the patterns of edge connections in the original graph, while preserving consistent behavior in downstream tasks. To achieve this, we first learn a mapping $f_\theta$ on original graph.The input to this mapping is the concatenated feature vector of a node pair $z_{uv}=[x_u;x_v]\in\mathbb{R}^{2d}$ and the output is the probability of the corresponding edge $p_{uv}=f_{\theta}(z_{uv})$. We then apply this mapping to the synthetic nodes to obtain the topology of the condensed graph\cite{DisCo}. By training on original graph, the edge predictor learns which combinations of node features tend to form edges, effectively capturing a transferable relationship between node and graph topology.

After obtaining the pre-trained edge predictor $f_\theta$, we perform all-pairs inference on the set of synthetic node features $X'$ to construct the condensed graph structure. For each node pair $\left(i,j\right)$, we compute
\begin{equation}A'_{ij}=f_\theta([x'_i;x'_j])\end{equation}
to obtain a probabilistic adjacency matrix $A'\in\mathbb{R}^{n\times n}$, where each element represents the confidence of edge existing between the corresponding node pair. Since the all-pairs matrix produces a dense graph, we further sparsify $A'$ by selecting a high-quantile threshold $\tau$ based on the distribution of all edge probabilities\cite{a50}. Only connections with probabilities significantly above random levels are retained, while entries below the threshold are set to 0:
\begin{equation}A'_{ij}=\begin{cases}A'_{ij}, & A'_{ij}\geq\tau, \\0, & \text{otherwise.}  \end{cases}\end{equation}

Finally, we extract the nonzero entries of the sparsified matrix $A'$ as the edge set and edge weights of the condensed graph, thereby obtaining the condensed graph structure. By leveraging the pre-trained edge predictor, we treat structure generation as a learnable functional mapping, eliminating the need for the original graph's adjacency structure during node synthesis. As long as the synthetic features preserve similarity to original graph features, the edge predictor can reconstruct a reasonable topology in the new feature space.

\section{EXPERIMENTS}
In this section, we conduct a series of experiments to evaluate the effectiveness of the proposed framework SP-ESGC. We first examine the expressive capacity of the condensed graphs by comparing the performance of node classification models trained on them. We then analyze the efficiency of SP-ESGC against other condensation methods, together with its generalization ability across models and datasets. In addition, we perform ablation studies to better understand the contribution of each component in our framework. All experiments are conducted on an NVIDIA GeForce RTX 3050 GPU.

% \begin{table}[htbp]
% \centering
% \caption{The statistics of five datasets.}
% \label{tab:table1}
% \begin{tabular}{@{}ccccccc@{}}
% \toprule
% Dataset     & \#Nodes & \#Edges    & \#Classes & \#Features  \\ \midrule
% Cora        & 2,708   & 5,429      & 7         & 1,433       \\
% Citeseer    & 3,327   & 4,732      & 6         & 3,703       \\
% Ogbn-arxiv  & 169,343 & 1,166,243  & 40        & 128         \\
% Flickr      & 89,250  & 899,756    & 7         & 500         \\
% Reddit      & 232,965 & 57,307,946 & 210       & 602         \\ \bottomrule
% \end{tabular}
% \end{table}

\begin{table*}[htbp]
\centering
\caption{The statistics of five datasets.}
\label{tab:table1}
\begin{tabular}{@{}ccccccc@{}}
\toprule
Dataset    & Task Type    & \#Nodes & \#Edges    & \#Classes & \#Features & Training/Validation/Test \\ \midrule
Cora       & Transductive & 2,708   & 5,429      & 7         & 1,433      & 140/500/1000             \\
Citeseer   & Transductive & 3,327   & 4,732      & 6         & 3,703      & 120/500/1000             \\
Ogbn-arxiv & Transductive & 169,343 & 1,166,243  & 40        & 128        & 90,941/29,799/48,603     \\
Flickr     & Inductive    & 89,250  & 899,756    & 7         & 500        & 44,625/22312/22313       \\
Reddit     & Inductive    & 232,965 & 57,307,946 & 210       & 602        & 15,3932/23,699/55,334    \\ \bottomrule
\end{tabular}
\end{table*}

\subsection{Experimental Settings}
We evaluate the condensation performance of SP-ESGC on five datasets, among which three are transductive datasets like Cora, Citeseer and Ogbn-arxiv, two are inductive datasets like Flickr and Reddit. For all datasets, we adopt the publicly available splits and experimental settings to ensure fair comparison. Detailed statistics of the datasets are reported in Table \ref{tab:table1}. For each dataset, we consider three condensation ratios ($r$). Specifically, $r$ is defined as the ratio between the number of condensed nodes $n$ and the number of original nodes $N$. For transductive datasets, $N$ denotes the total number of nodes in the entire graph, whereas for inductive datasets, $N$ refers to the number of nodes in the training subgraph of the graph.We compared the proposed method with several baselines: (1) three coreset methods including Random, Herding\cite{Herding} and K-Center\cite{K-Center}; (2) five graph condensation methods including GCOND\cite{GCond}, SGDD\cite{SGDD}, SimGC\cite{SimGC}, GC-SNTK\cite{GC-SNTK} and SFGC\cite{SFGC}.

\begin{table*}[htbp]
\centering
\caption{Node classification performance under different graph condensation ratios on different graph condensation methods.Performance is reported as test accuracy (\%), with "$\pm$" the standard deviation across three trials.Best results are in bold and the second-bests are underlined.OOM means out-of-memory.}
\label{tab:table2}
\begin{tabular}{@{}cccccccccccc@{}}
\toprule
Dataset  & Ratio (r) & Random  & Herding  & K-Center  & Gcond  & SGDD  & SimGC  & GC-SNTK  & SFGC  & SP-ESGC  & Whole  Dataset            \\ \midrule
\multirow{3}{*}{Cora}  
& 1.30\%  & 63.6±3.7  & 67.0±1.3  & 64.0±2.3  & 79.8±1.3  & 80.1±0.7  & 80.8±2.3  & {\ul 81.3±0.2} & 80.1±0.4  & \textbf{82.6±0.6} & \multirow{3}{*}{81.2±0.2} \\
& 2.60\%  & 72.8±1.1  & 73.4±1.0  & 73.2±1.2  & 80.1±0.6  & 80.6±0.8  & 80.9±2.6  & 81.5±0.7  & {\ul 81.7±0.5}  & \textbf{82.7±0.1} &  \\
& 5.20\%  & 76.8±0.1  & 76.8±0.1  & 76.7±0.1  & 79.3±0.3  & 80.4±1.6  & {\ul 82.1±1.3}  & 81.3±0.2  & 81.6±0.8  & \textbf{82.5±0.4} &  \\ \midrule
\multirow{3}{*}{Citeseer}   
& 0.90\%  & 54.4±4.4  & 57.1±1.5  & 52.4±2.8  & 70.5±1.2  & 69.5±0.4  & \textbf{73.8±2.5}  & 66.4±1.0  & 71.4±0.5  & {\ul 73.5±0.1}  & \multirow{3}{*}{71.7±0.1} \\
& 1.80\%  & 64.2±1.7  & 66.7±1.0  & 64.3±1.0  & 70.6±0.9  & 70.2±0.8  & 72.2±0.5  & 68.4±1.1  & {\ul 72.4±0.4}  & \textbf{72.8±0.2} &                           \\
& 3.60\%  & 69.1±0.1  & 69.0±0.1  & 69.1±0.1  & 69.8±1.4  & 70.3±1.7  & {\ul 71.1±2.8}  & 69.8±0.8  & 70.6±0.7  & \textbf{72.0±0.4} &  \\ \midrule
\multirow{3}{*}{Ogbn-arxiv} 
& 0.05\%  & 47.1±3.9  & 52.4±1.8  & 47.2±3.0  & 59.2±1.1  & 60.8±1.3  & 63.6±0.8  & {\ul 64.4±0.2}  & \textbf{65.5±0.7}  & 64.3±0.4  & \multirow{3}{*}{71.4±0.1} \\
& 0.25\%  & 57.3±1.1  & 58.6±1.2  & 56.8±0.8  & 63.2±0.3  & 65.8±1.2  & {\ul 66.4±0.3}  & 65.1±0.8  & 66.1±0.4  & \textbf{66.7±0.1} &  \\
& 0.50\%  & 60.0±0.9  & 60.4±0.8  & 60.3±0.4  & 64.0±0.4  & {\ul 66.3±0.7}  &  \textbf{66.8±0.4}  & 65.4±0.5  & \textbf{66.8±0.4}  & \textbf{66.8±0.1}  &  \\ \midrule
\multirow{3}{*}{Flickr}     
& 0.10\%  & 41.8±2.0  & 42.5±1.8  & 42.0±0.7  & 46.5±0.4  & 46.5±0.1  & 45.3±0.7  & {\ul 46.7±0.1}  & 46.6±0.2  & \textbf{47.2±0.2}  & \multirow{3}{*}{47.2±0.1}  \\
& 0.50\%  & 44.0±0.4  & 43.9±0.9  & 43.2±0.1  & 45.2±0.3  & 46.4±0.2  & 45.6±0.4  & 46.8±0.1  & {\ul 47.0±0.1}  & \textbf{47.2±0.3} &  \\
& 1.00\%  & 44.6±0.2  & 44.4±0.6  & 44.1±0.4  & \textbf{47.1±0.1}  & 46.3±0.1 & 43.8±1.5  & {\ul 46.5±0.2}  & \textbf{47.1±0.1}  & \textbf{47.1±0.3} &  \\ \midrule
\multirow{3}{*}{Reddit}     
& 0.05\%  & 46.1±4.4  & 53.1±2.5  & 46.6±2.3  & 88.0±1.8  & {\ul 90.3±0.1}  & 89.6±0.6  & OOM  & 89.7±0.2  & \textbf{90.7±0.0}  & \multirow{3}{*}{93.9±0.0}  \\
& 0.10\%  & 58.0±2.2  & 62.7±1.0  & 53.0±3.3  & 89.6±0.7  & {\ul 90.7±0.1}  & 90.6±0.3  & OOM  & 90.0±0.3  & \textbf{91.6±0.0}  &  \\
& 0.20\%  & 66.3±1.9  & 71.0±1.6  & 58.5±2.1  & 90.1±0.5  & 91.3±0.2  & {\ul 91.4±0.2}  & OOM  & 89.9±0.4  & \textbf{92.3±0.0} &  \\ \bottomrule
\end{tabular}
\end{table*}

\subsection{Prediction Accuracy}
We train node classification models on the condensed graphs and predict labels on the test sets. Classification accuracy is used as the evaluation metric to measure the representational capacity of the condensed graphs. We evaluate the representation quality of the condensed graphs under three different condensation ratios and report the test accuracy of all methods on different datasets in Table \ref{tab:table2}. From the results, we observe that coreset methods perform worse than approaches specifically designed for graph condensation. This gap becomes more evident as the condensation ratio decreases. These observations further underline the importance of condensation methods to graph data. SP-ESGC achieves the best or second-best performance in the majority of cases.Specifically, under low-ratio settings such as Cora ($r$=1.30\%), Flickr ($r$=0.10\%), and Reddit ($r$=0.05\%), SP-ESGC consistently achieves the best results, indicating that it can effectively preserve both structural and semantic information even at low condensation ratio. In contrast, GC-SNTK encounters out-of-memory (OOM) issues on Reddit, revealing scalability limitations on large-scale graph datasets, whereas SP-ESGC maintains high efficiency.

\subsection{Condensation Time}
In this section, we compare the condensation time of SP-ESGC with that of five graph condensation baselines. To ensure a fair evaluation, we separately record the time spent on graph condensation, edge predictor model pretraining and condensed graph performance evaluation for SP-ESGC. The total runtime is added these time together and then compared against other methods. As shown in Table \ref{tab:table3}, SP-ESGC is shown to be significantly faster than existing graph condensation methods. Notably, on the Reddit dataset, the condensation time of SP-ESGC is reduced to approximately one-sixteenth of that required by the second fastest baseline. This result highlights the efficiency advantage of SP-ESGC, especially in large-scale graph.

\begin{table}[htbp]
\centering
\caption{The condensation time (seconds) of our proposed SP-ESGC and baselines on five datasets. The r of the five datasets are respectively set to 2.60\%, 1.80\%, 0.25\%, 0.50\% and 0.10\%.}
\label{tab:table3}
\begin{tabular}{cccccc}
% \hline
\toprule
\multicolumn{1}{l}{} & Cora          & Citeseer      & Ogbn-arxiv     & Flickr        & Reddit         \\ \midrule
GCond                & 653.9         & 940.3         & 13521.7        & 1455.6        & 20528.8        \\
SGDD                 & 4848.6        & 3091.2        & 35179.7        & 26767.4       & 378220.9       \\
SimGC                & 289.3         & 495.7         & 476.6          & 744.5         & 2655.9         \\
GC-SNTK              & 92.1          & 69.7          & 28897.8        & 889.8         & OOM            \\
SFGC                 & 5895.8        & 3807.2        & 156975.2       & 46706.7       & 370089.0         \\
SP-ESGC                 & \textbf{12.4} & \textbf{26.4} & \textbf{143.4} & \textbf{22.6} & \textbf{162.5} \\ 
% \hline
\bottomrule
\end{tabular}
\end{table}

\subsection{Generalization Ability}
\begin{figure*}[htbp]
\centering
\includegraphics[width=1.0\textwidth]{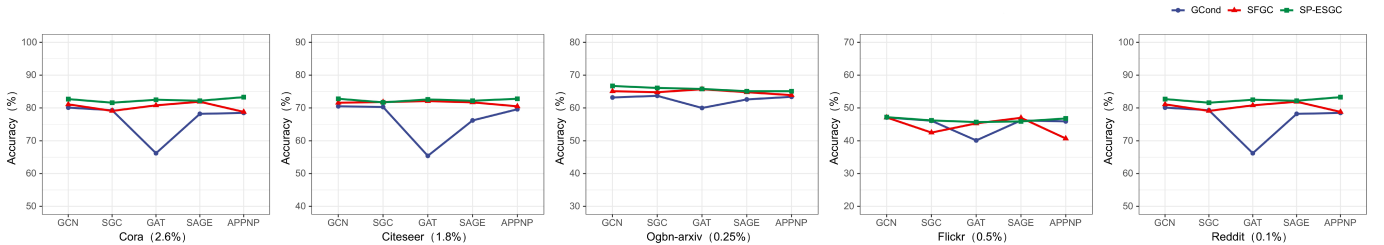}
\caption{The generalization capabilities of SP-ESGC on five datasets.}
\label{fig2}
\end{figure*}

In this section, we evaluate the generalization ability of different graph condensation methods by training various GNN architectures on the condensed graphs. Specifically, we consider GCN\cite{GCN}, SGC\cite{SGC}, GAT\cite{GAT}, GraphSAGE\cite{GraphSAGE}, APPNP\cite{APPNP}, and compare SP-ESGC with two representative graph condensation methods, GCond and SFGC. As shown in Fig.\ref{fig2}, SP-ESGC demonstrates consistently strong performance across diverse GNN architectures and datasets. The performance of GCond exhibits noticeable degradation on certain architectures (e.g. GAT), while SP-ESGC maintains relatively stable accuracy with smaller fluctuations across GNNs. The results suggest that SP-ESGC provides better generalization and stability.

\subsection{Ablation Study}
To analyze the contribution of each component in SP-ESGC, we conduct ablation studies on multiple datasets with three variants. In "w/o HKP", we remove the heat kernel feature propagation module and directly use the original node features for node condensation. In "w/o EP", the pre-trained edge predictor is removed, edges between condensed nodes are constructed based on cosine similarity of node features. In "w K-means", we replace the proposed class condensation strategy with K-means clustering. Results are reported in Table \ref{tab:table5}. We observe that the full model consistently achieves the best performance across all datasets, indicating that each component contributes to the effectiveness of SP-ESGC. The ablation results clearly show that: (1) Heat kernel feature propagation introduces global information prior to condensation, providing a more stable node representation for subsequent clustering and structure generation. (2) The method based on feature similarity are insufficient for generating meaningful graph structures. The edge predictor is able to capture connection rules for edges from the original graph and transfer them to the condensed nodes. (3) Simple Euclidean-space clustering fails to capture the intrinsic structure of node representations. In contrast, intra-class spectral space clustering more accurately models the nonlinear distributions, producing more representative node centroids.

\begin{table}[htbp]
\centering
\caption{The ablation study of SP-ESGC.}
\small
\setlength{\tabcolsep}{2pt}
\label{tab:table5}
\begin{tabular}{@{}*{6}{c}@{}}
\toprule
Method & Cora & Citeseer & Ogbn-arxiv & Flickr & Reddit \\ 
\midrule
% ESGC w/o HKP   & 78.6±0.3 & 70.4±0.2 & 65.4±0.3 & 46.7±0.4 & 90.5±0.1 \\
% ESGC w/o EP   & 81.9±0.2 & 72.5±0.2 & 66.0±0.2 & 45.8±0.1 & 91.1±0.1 \\
% ESGC w K-means & 81.4±0.2 & 72.0±0.2 & 66.6±0.0 & 46.7±0.2 & 91.4±0.0 \\
% ESGC           & \textbf{82.7±0.1} & \textbf{72.8±0.2} & \textbf{66.7±0.1} & \textbf{47.2±0.3} & \textbf{91.6±0.0} \\ 
SP-ESGC w/o HKP   & 78.6$_{\text{\tiny ±0.3}}$ & 70.4$_{\text{\tiny ±0.2}}$ & 65.4$_{\text{\tiny ±0.3}}$ & 46.7$_{\text{\tiny ±0.4}}$ & 90.5$_{\text{\tiny ±0.1}}$ \\
SP-ESGC w/o EP    & 81.9$_{\text{\tiny ±0.2}}$ & 72.5$_{\text{\tiny ±0.2}}$ & 66.0$_{\text{\tiny ±0.2}}$ & 45.8$_{\text{\tiny ±0.1}}$ & 91.1$_{\text{\tiny ±0.1}}$ \\
SP-ESGC w K-means & 81.4$_{\text{\tiny ±0.2}}$ & 72.0$_{\text{\tiny ±0.2}}$ & 66.6$_{\text{\tiny ±0.0}}$ & 46.7$_{\text{\tiny ±0.2}}$ & 91.4$_{\text{\tiny ±0.0}}$ \\
SP-ESGC           & \textbf{82.7$_{\text{\tiny ±0.1}}$} & \textbf{72.8$_{\text{\tiny ±0.2}}$} & \textbf{66.7$_{\text{\tiny ±0.1}}$} & \textbf{47.2$_{\text{\tiny ±0.3}}$} & \textbf{91.6$_{\text{\tiny ±0.0}}$} \\ 
\bottomrule
\end{tabular}
\end{table}

\section{CONCLUSION}
In this work, we study graph condensation from an efficiency and structure-aware perspective and propose a simple and effective framework, SP-ESGC. By decoupling node condensation from graph structure generation, SP-ESGC avoids complex bi-level optimization and reduces computational overhead. The proposed heat kernel feature propagation introduces global information in a stable and controllable manner, while the clustering strategy preserves intra-class discriminability under condensation. Moreover, the edge predictor enables to generate flexible and scalable topology construction. Experiments on both transductive and inductive datasets show that SP-ESGC achieves strong performance across various condensation ratios and GNN architectures. Overall, SP-ESGC offers an efficient solution for graph condensation, making it well suited for large-scale and resource-constrained scenarios.

% \begin{table*}[htbp]
% \centering
% \caption{The statistics of five datasets.}
% \label{tab:table1}
% \begin{tabular}{@{}ccccccc@{}}
% \toprule
% Dataset    & Task Type    & \#Nodes & \#Edges    & \#Classes & \#Features & Training/Validation/Test \\ \midrule
% Cora       & Transductive & 2,708   & 5,429      & 7         & 1,433      & 140/500/1000             \\
% Citeseer   & Transductive & 3,327   & 4,732      & 6         & 3,703      & 120/500/1000             \\
% Ogbn-arxiv & Transductive & 169,343 & 1,166,243  & 40        & 128        & 90,941/29,799/48,603     \\ \midrule
% Flickr     & Inductive    & 89,250  & 899,756    & 7         & 500        & 44,625/22312/22313       \\
% Reddit     & Inductive    & 232,965 & 57,307,946 & 210       & 602        & 15,3932/23,699/55,334    \\ \bottomrule
% \end{tabular}
% \end{table*}

\end{document}